\documentclass[runningheads]{llncs}
\usepackage[T1]{fontenc}
\usepackage{graphicx}
\usepackage{booktabs}
\usepackage{siunitx}
\usepackage{adjustbox}
\usepackage{tabularx}
\usepackage{pifont}
\usepackage{array}
\usepackage{xcolor}
\usepackage{url}
\usepackage{caption}
\usepackage{multirow}
\usepackage[acronym,nonumberlist]{glossaries}
\usepackage{float}
\usepackage{tikz}
\usetikzlibrary{
    positioning,
    arrows.meta,
    shapes.geometric,
    calc
}
\usepackage{tcolorbox}
\usepackage{listings}
\begin{document}
\title{Bayesian Uncertainty Propagation for Agentic RAG Pipelines: A Proof-of-Concept Study on Multi-Hop Question Answering}
\titlerunning{Uncertainty Monitoring for LLM Agents}
%
\author{Louis Donaldson\inst{1}\orcidID{0009-0000-3862-8061} \and
Connor Walker\inst{1}\orcidID{0009-0008-8181-1644} \and
Koorosh Aslansefat\inst{1}\orcidID{0000-0001-9318-8177} 
\and 
Yiannis Papadopoulos\inst{1}\orcidID{0000-0001-7007-5153}}

\authorrunning{L. Donaldson et al.}
%
\institute{University of Hull, Hull HU6 7RX, UK \\
\email{l.donaldson-2020@hull.ac.uk}}
\maketitle              
\begin{abstract}
Trustworthy deployment of Agentic Retrieval-Augmented Generation (RAG) systems requires mechanisms for estimating when multi-stage reasoning pipelines may fail. This paper presents an uncertainty-aware Agentic Retrieval-Augmented Generation (RAG) framework in which planner, evaluator and generator stages produce uncertainty signals derived from semantic divergence and generator self-evaluation. These signals are propagated through a Bayesian Network (BN) to estimate system-level uncertainty and provide node-level indicators of potential failure points across the workflow. The approach is evaluated on StrategyQA and HotpotQA using GPT-3.5-Turbo and GPT-4.1-Nano, with Area Under the Receiver Operating Characteristic Curve (AUROC), Area Under the Accuracy-Rejection Curve (AUARC), Expected Calibration Error (ECE), and Brier Score used to assess discrimination, selective prediction and calibration. Results show that Bayesian propagation is more effective on HotpotQA, where uncertainty accumulates across multi-hop reasoning stages, while StrategyQA exposes limitations caused by miscalibration and unreliable upstream signals. The study positions Bayesian uncertainty propagation as a promising but preliminary mechanism for monitoring Agentic RAG systems, with future validation required in industrial domains such as Offshore Wind (OSW) maintenance decision support.

 \and AI-Enabled Industrial Decision Support Systems
\end{abstract}

\newacronym{rag}{RAG}{Retrieval-Augmented Generation}
\newacronym{llm}{LLM}{Large Language Model}
\newacronym{bn}{BN}{Bayesian Network}
\newacronym{cpt}{CPT}{Conditional Probability Table}
\newacronym{eu}{EU}{Extrinsic Uncertainty}
\newacronym{iu}{IU}{Intrinsic Uncertainty}
\newacronym{auroc}{AUROC}{Area Under the Receiver Operating Characteristic Curve}
\newacronym{auarc}{AUARC}{Area Under the Accuracy-Rejection Curve}
\newacronym{ece}{ECE}{Expected Calibration Error}
\newacronym{pmi}{PMI}{Pointwise Mutual Information}
\newacronym{pomdp}{POMDP}{Partially Observable Markov Decision Process}
\newacronym{osw}{OSW}{Offshore Wind}
\newacronym{maa}{MAA}{Multi-Agent Architecture}
\newacronym{pdm}{PdM}{Predictive Maintenance}
\newacronym{se}{SE}{Semantic Entropy}
\newacronym{pca}{PCA}{Principal Component Analysis}
\newacronym{mas}{MAS}{Multi-Agent System}
\newacronym{json}{JSON}{JavaScript Object Notation}
\newacronym{react}{ReAct}{Reasoning and Acting}
\newacronym{sd}{SD}{Semantic Divergence}


\section{Introduction}
As \glspl{llm} become increasingly integrated into industrial workflows, their ability to automate routine tasks and provide human-understandable interpretations of complex data has the potential to improve operational efficiency across a wide range of sectors. As confidence in these systems grows, they are expected to play increasingly important roles in business-critical processes. One potential application area is \gls{osw} maintenance scheduling, where increasing turbine complexity and expanding wind farm capacity require the interpretation of large volumes of operational data to support maintenance planning. Recent work has demonstrated the potential of \gls{llm}-based decision-support and safety monitoring within this domain, while also highlighting the need for reliable run-time monitoring and trustworthy uncertainty estimation \cite{walker_2024_safellm}.

\subsection{Background and Motivation}
The adoption of \glspl{llm} within specialised domains is accompanied by significant challenges regarding transparency, reliability and trust. In healthcare, the inability to inspect the reasoning process behind generated recommendations may affect clinical decision making and patient safety, while privacy considerations further complicate deployment \cite{powell_2025_generating,bilal_2025_llms}. Similar concerns arise in finance and legal applications, where opaque reasoning makes regulatory compliance and decision auditing difficult \cite{bilal_2025_llms,powell_2025_generating}. In scheduling, manufacturing and \gls{pdm}, operators often require understandable explanations before trusting automated decisions, yet conventional black-box models provide limited insight into the reasoning behind their outputs \cite{powell_2025_generating,zhang_2026_dschellm,hughes_2025_costefficiency}.

\gls{rag} partially addresses these limitations by grounding generated responses in retrieved knowledge sources, allowing supporting evidence to be inspected and verified by human operators \cite{deng_2024_from,dogra_2025_generative,walker_2025_raguard}. Hybrid and \glspl{maa} further improve reasoning by decomposing complex tasks across specialised agents or combining \gls{llm} reasoning with domain-specific models, producing more interpretable reasoning workflows \cite{harbola_2025_prescriptive,yuan_2025_towards}. In parallel, prompting paradigms such as ReAct and context engineering improve reasoning consistency through external tool use and structured contextual guidance \cite{bilal_2025_llms,venkatachalam_2025_integrating}.

Although these approaches improve reasoning quality and interpretability, they do not explicitly model how uncertainty propagates across sequential reasoning stages. In multi-agent \gls{rag} pipelines, uncertainty introduced during planning, retrieval or evaluation may influence downstream decisions and ultimately affect the reliability of the final response. This motivates the development of lightweight uncertainty propagation methods capable of monitoring confidence throughout the reasoning workflow. The contributions of this paper are threefold: 

\begin{enumerate}
    \item We propose a lightweight uncertainty-aware monitoring architecture for Agentic \gls{rag} systems, where Planner, Evaluator and Generator agents produce complementary uncertainty signals during multi-stage reasoning.
    \item We introduce a \gls{bn}-based uncertainty propagation model that combines node-level uncertainty estimates into an interpretable system-level confidence estimate capable of indicating which stages contribute most to system-level uncertainty.
    \item We present a proof-of-concept empirical evaluation on the StrategyQA and HotpotQA multi-hop question answering benchmarks using GPT-3.5-Turbo and GPT-4.1-Nano, demonstrating the strengths and limitations of Bayesian uncertainty propagation across multiple uncertainty evaluation metrics, including \gls{auroc}, \gls{auarc}, \gls{ece} and Brier Score.
\end{enumerate}
\color{black}
\section{Methodology on Uncertainty Quantification}
Agentic \gls{rag} systems operate as sequential decision-making processes that can be modelled as a \gls{pomdp} \cite{mishra_2026_sok}. Because each stage depends on previous decisions, uncertainty introduced during planning, retrieval or evaluation may propagate through the workflow, leading to cascading errors and semantic drift. Consequently, reliable uncertainty quantification is required to monitor confidence throughout multi-stage reasoning.

\subsection{Token-level Entropy}
Also referred to as naive entropy, it represents the uncertainty of a \gls{llm} at the level of individual word-pieces or characters during the text generation process \cite{farquhar_2024_detecting}. In autoregressive models, this uncertainty is derived from the softmax probability distribution over all possible next tokens in the model's vocabulary. Given an input $x$ and model parameters $\theta$, the predictive distributions of the output $y$ can be written as $p_{\theta}(y | x)$, and for a generated sequence of outputs $y = (z_1,...,z_L)$ with length $L$, the probability can be decomposed into conditional token probabilities:
 \begin{equation}
     p_{\theta}(y \mid x) = \prod_{t=1}^{L} p_{\theta}(z_t \mid z_{<t}, x)
 \end{equation}
 Taking the logarithm to yield the log probability of the sequence and applying length normalisation to ensure a fair comparison between different response lengths, this can be computed as:
 \begin{equation}
     s_i = \frac{1}{L_i} \sum_{t=1}^{L_i} \log p_{\theta}(z_t \mid z_{<t}, x)
 \end{equation}
which is then transformed into a feature of confidence denoted as $q_i = exp(s_i)$. This allows the probability of uncertainty to be estimated as $$P(X_i = success | q_i) = f_\textbf{cal}(q_i)$$
 
where $f_\textbf{cal}$ denotes a calibration function estimated using runs. This value is then incorporated into the \gls{bn} as probabilistic evidence.

Although naive entropy metrics are not direct indicators of the truthfulness of model outputs, in multi-agent systems, they can be used to parameterise the Prior Probabilities of nodes in a \gls{bn}. They can act as local confidence estimates that are transformed into bounded features and then propagated through the \gls{bn} to understand how the uncertainty accumulates across workflow stages.

The use of logprobs in this system is utilised for the Generator node using the P(True) self-evaluation approach. This is a supervised "in-context" uncertainty estimation to evaluate the truthfulness of its own outputs \cite{farquhar_2024_detecting}. Once the \gls{rag} pipeline has finished executing and produced an answer, it is then posed a further question "Are you sure that the answer provided is correct?" to which it can only reply with a single token (true/false or yes/no). The confidence score, or P(True) is the probability that the model assigns to the affirmative token \cite{farquhar_2024_detecting}.

\subsection{Semantic Divergence}
Also referred to as "semantic drift", can occur during iterative retrieval processes. As an agent autonomously reformulates its search queries across multiple steps, it can gradually diverge from the original information requirement \cite{mishra_2026_sok}. Because Agentic \gls{rag} systems utilise a sequential decision process, minor deviation in early steps can compound across iterations, leading to "reasoning collapse" and unnecessary computational costs \cite{mishra_2026_sok}. Semantic divergence can be used as a metric to identify confabulations; hallucinations that are both arbitrary and semantically misaligned with the input context \cite{halperin_2025_promptresponse}.

\subsubsection{Theoretical Foundations}
The Semantic Divergence formulation in this paper is motivated by the UProp framework of Duan et al,. \cite{duan_2025_uprop}, which decomposes sequential decision uncertainty into \gls{iu} and \gls{eu}. In UProp, \gls{eu} at step $t$ is estimated via the \gls{pmi} between the current step's decision distribution and each preceding step's committed decision:

\begin{equation}
    \widehat{\text{PMI}}(y_t;\, y^{(k)}_{t-1} \mid x)
    \;=\;
    -\log \sum_{n=1}^{N} K_N\!\left(d\!\left(y^{(n)}_{t-1},\, y^{(k)}_{t-1}\right)\right)
    \label{eq:uprop_pmi}
\end{equation}

\noindent where $K_N$ is a Gaussian kernel applied to the fuzzy string distance $d$ between the $N$ samples at step $t{-}1$ and the committed decision $y^{(k)}_{t-1}$. Semantic Divergence instantiates an analogous measurement within each node, replacing the Gaussian kernel over string distances with the Wasserstein distance between the committed output distribution and the re-sampled output distribution in a projected embedding space. Both metrics measure the spread of the decision distribution around the committed trajectory point; Semantic Divergence therefore provides a per-node \gls{iu} estimate that is directly comparable to UProp's PMI-based \gls{eu} accumulation.

\subsubsection{Comparison with Semantic Entropy}
\gls{se} \cite{kuhn_2023_semantic_entropy} clusters $K$ samples by semantic equivalence and computes Shannon entropy over the resulting cluster distribution:

\begin{equation}
    \text{SE} \;=\; -\sum_{c} p(c) \log p(c)
    \label{eq:se}
\end{equation}

\gls{se} measures uncertainty by clustering multiple sampled outputs according to semantic equivalence and computing the entropy of the resulting distribution. To prevent repeated \gls{json} outputs collapsing into a single \gls{se} cluster, this paper proposes the use of semantic divergence to anchor measurements to the committed output rather than measuring within-sample spread. 

\subsection{Bayesian Networks}
In this \gls{mas} approach, \glspl{bn} are used to represent uncertainty propagation by modelling the dependencies between agents. Each stage of the workflow is modelled as a node, and the dependencies of each node are captured through directed edges. A combination of semantic-divergence scores and token-level log-probabilities are used to quantify uncertainty at each node, which is then propagated through the \gls{bn} to quantify the overall uncertainty level.


\glspl{bn} are probabilistic graphical models that represent conditional dependencies between random variables using directed acyclic graphs \cite{liu_2019_a}. Let $X=\{ X_1,..., X_n\}$ denote the workflow tasks, where $X_1$, $X_2$ and $X_3$ correspond to the Planner, Evaluator and Generator agents respectively. 
\color{black}
\begin{equation}
P(X_1,...,X_n)=\prod_{i=1}^{n}P(X_i|\mathrm{Pa}(X_i))
\end{equation}

where $Pa(X_i)$ denotes the set of parent nodes of $X_i$. The corresponding \gls{cpt} specifies the conditional probability of each node given its parents. Within the proposed Agentic \gls{rag} framework, each node represents the uncertainty state of the Agentic \gls{rag} stage, with confidence scores parameterising the prior probability according to $P=(X_i=success)=c(y_i)$.
\begin{figure}
    \centering    \includegraphics[width=0.75\linewidth]{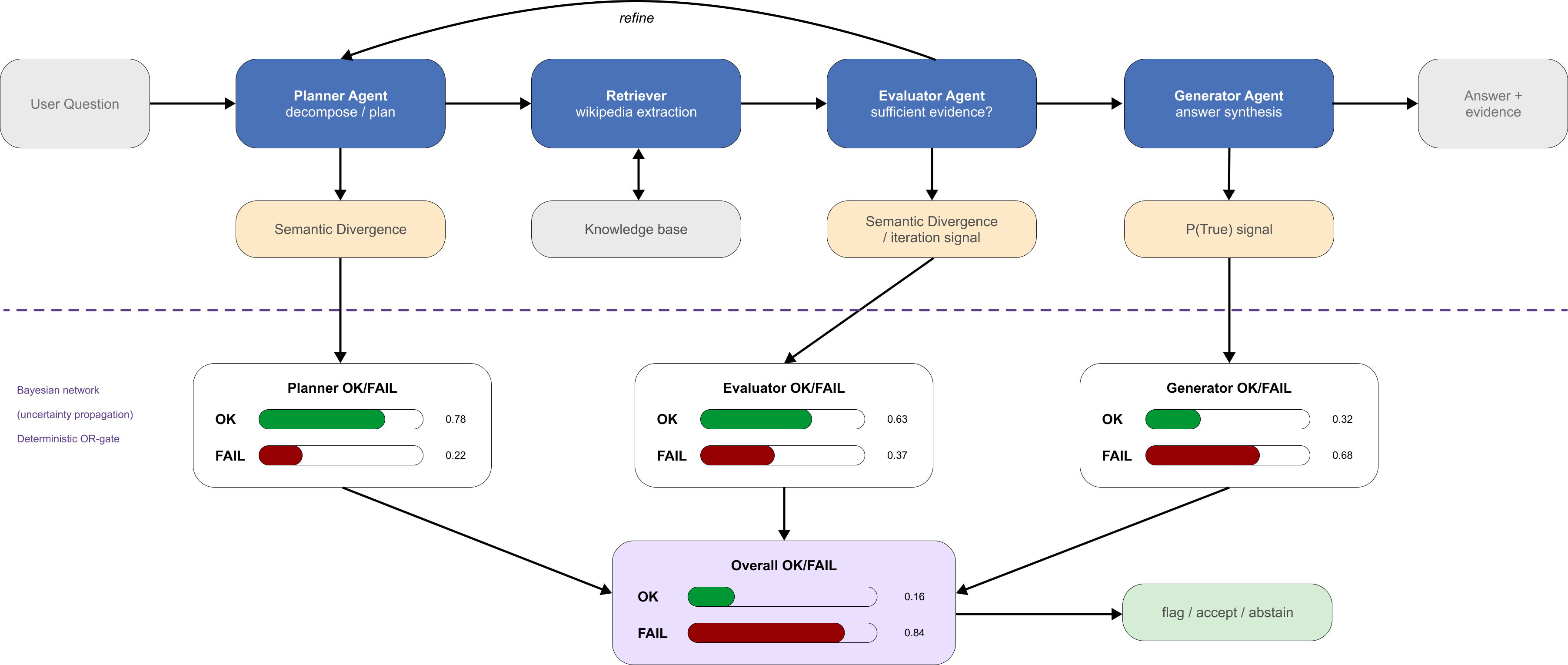}
    \caption{Agentic \gls{rag} pipeline with \gls{bn} uncertainty propagation used to flag/accept/abstain answers based on Overall OK/FAIL.}
\label{fig:pipeline}
\end{figure}

The \gls{bn} representation used in this study represents an abstracted execution trace of the workflow rather than the full control logic of the Agentic system. Temporal modelling does not extend further than parent-child dependencies, nor is there any task priority or repeated executions, feedback loops or iterative cycles. If the Evaluator deems there is insufficient evidence to produce a confident answer, the final accepted cycle is used as evidence for the corresponding node. 

For the Overall node that relies on preceding steps in the workflow, the confidence scores contribute to the \gls{cpt}. As seen in Table \ref{tab:or_gate_cpt} the relationship between Overall and its parent nodes is characterised for this study as a deterministic OR gate. This implies that Overall is FAIL when one node in the pipeline is uncertain or FAIL. The threshold for determining a FAIL or OK in this study is set at 0.5 (or 50 \%). If an uncertainty value of a node exceeds this, it is considered a FAIL. Using a deterministic OR gate in this study is intended to test if the simplification of the uncertainty propagation behaviour throughout an Agentic \gls{rag} pipeline is a viable method to quantify the overall uncertainty of responses and lead to detection of possible hallucinations or confabulations.

\begin{table}[tb]
\centering
\caption{\gls{cpt} for \textit{Overall} under a deterministic OR gate.}
\label{tab:or_gate_cpt}
\renewcommand{\arraystretch}{1.1}
\begin{tabular}{ccc|c}
\toprule
\textbf{Planner}& \textbf{Evaluator}& \textbf{Generator}& $P(\text{Overall} = \text{OK})$ \\
\midrule
OK   & OK   & OK   & 1.0 \\
OK   & OK   & FAIL & 0.0 \\
OK   & FAIL & OK   & 0.0 \\
OK   & FAIL & FAIL & 0.0 \\
FAIL & OK   & OK   & 0.0 \\
FAIL & OK   & FAIL & 0.0 \\
FAIL & FAIL & OK   & 0.0 \\
FAIL & FAIL & FAIL & 0.0 \\
\bottomrule
\end{tabular}
\end{table}

New evidence is also included in the network and the posterior probabilities are updated alongside the workflow run. The interpretability of the \glspl{bn} graphical structure makes it possible to identify which components can contribute most to the overall uncertainty.

The objective of performing probabilistic inference, once the \gls{bn} structure and \glspl{cpt} are defined, is to estimate posterior probabilities across the workflow. The objective of this is to determine how uncertainty in each node affects the reliability of the final workflow outcome.

\section{Experiments \& Results}
In this section, we provide the experimental setup and discuss the performance of multi-step reasoning benchmarks including StrategyQA \cite{geva_2021_did} and HotpotQA \cite{yang_2018_hotpotqa}. These multi-hop, question-answering benchmarks were chosen due to their popularity and academic presence. In these, \glspl{llm} are tasked with answering questions that require multi-hop reasoning where each action will provide a keyword or phrase request to a Wikipedia engine for retrieval using a \gls{react} \cite{yao_2023_react} methodology. Both benchmark experiments are run with 200 randomly sampled questions.

In this study we use commercial \glspl{llm} provided by OpenAI, GPT-3.5-Turbo and GPT-4.1-Nano. For the generative hyper-parameters, we assign the Planner and Generator agents a temperature of 0.2, the Evaluator a temperature of 0.8 with a maximum number of new tokens as 256. By default the per-step sample number of $N$ is set to 10.

As an indicative comparison from the aforementioned benchmarks, we are using UProp's \cite{duan_2025_uprop} \gls{auroc} scores as an external reference for the results produced in the approach proposed in this study.

\begin{table}[tb]
\caption{\gls{auroc} results per node and \gls{bn} posterior (Overall) across benchmarks. Bold denotes the best \gls{bn} score per row. \gls{auroc} $>$ 0.5 indicates better-than-random discrimination.}
\centering
\footnotesize
\setlength{\tabcolsep}{0pt}
\begin{tabularx}{\textwidth}{lXXXXX}
\toprule
\multicolumn{6}{c}{\textbf{Benchmark: StrategyQA (200 Samples)}} \\
\midrule
\textbf{Model} & \centering\textbf{Planner} & \centering\textbf{Evaluator} & \centering\textbf{Generator} & \centering\textbf{BN (Ours)} & \centering\arraybackslash\textbf{UProp} \\
\midrule
GPT-3.5-Turbo  & \centering 0.4138 & \centering 0.4799 & \centering \textbf{0.5881} & \centering 0.5592 & \centering\arraybackslash 0.6040 \\
GPT-4.1-Nano   & \centering 0.4696 & \centering 0.4675 & \centering \textbf{0.6213} & \centering 0.6083 & \centering\arraybackslash 0.5440 \\
\midrule
\multicolumn{6}{c}{\textbf{Benchmark: HotpotQA (200 Samples)}} \\
\midrule
\textbf{Model} & \centering\textbf{Planner} & \centering\textbf{Evaluator} & \centering\textbf{Generator} & \centering\textbf{BN (Ours)} & \centering\arraybackslash\textbf{UProp} \\
\midrule
GPT-3.5-Turbo  & \centering 0.6209 & \centering 0.6117 & \centering 0.6993 & \centering \textbf{0.7160} & \centering\arraybackslash 0.7470 \\
GPT-4.1-Nano   & \centering 0.5919 & \centering 0.5369 & \centering 0.7145 & \centering \textbf{0.7240} & \centering\arraybackslash  0.7188 \\
\bottomrule
\end{tabularx}
\label{tab:auroc}
\end{table}

As shown in Table \ref{tab:auroc}, the \gls{auroc} results demonstrate contrasting behaviour across the two benchmarks. On StrategyQA, the Generator provides the strongest individual uncertainty signal, while the Planner and Evaluator perform close to random, limiting the effectiveness of Bayesian uncertainty propagation. Consequently, the Overall \gls{bn} remains below the UProp baseline for GPT-3.5-Turbo, although it exceeds the baseline for GPT-4.1-Nano. In contrast, all nodes achieve above random discrimination on HotpotQA, and the Overall \gls{bn} becomes competitive with UProp, exceeding the baseline for GPT-4.1-Nano while remaining only slightly below it for GPT-3.5-Turbo. These results suggest that Bayesian uncertainty propagation is most effective when uncertainty accumulates across multiple reasoning stages. 

\begin{table}[tb]
\centering
\caption{\gls{auarc} results across benchmarks. Higher values indicate greater improvement in accuracy under selective prediction.
Bold denotes the highest \gls{auarc} value in each row.}
\label{tab:auarc}
\renewcommand{\arraystretch}{1.1}
\setlength{\tabcolsep}{5pt}
\begin{tabular}{llcccc}
\toprule
\textbf{Benchmark} & \textbf{Model}
    & \textit{Planner}
    & \textit{Evaluator}
    & \textit{Generator}
    & \textbf{BN (Ours)} \\
\midrule
\multirow{2}{*}{StrategyQA}
    & GPT-3.5-Turbo & 0.4968 & 0.5447 & \textbf{0.6387} & 0.5952 \\
    & GPT-4.1-Nano  & 0.5554 & 0.5680 & \textbf{0.6831} & 0.6764 \\
\midrule
\multirow{2}{*}{HotpotQA}
    & GPT-3.5-Turbo & 0.3920 & 0.3611 & 0.4148 & \textbf{0.4254} \\
    & GPT-4.1-Nano  & 0.3213 & 0.2733 & 0.3669 & \textbf{0.3862} \\
\bottomrule \\
\end{tabular}
\end{table}

The \gls{auarc} results presented in Table \ref{tab:auarc} complement the \gls{auroc} analysis by evaluating selective prediction performance. On StrategyQA, the Generator provides the strongest individual uncertainty estimate, indicating that P(True) self-evaluation is sufficient for effective selective prediction on the simpler benchmark. In contrast, the Overall \gls{bn} achieves the highest \gls{auarc} on HotpotQA, suggesting that combining uncertainty across multiple pipeline stages becomes increasingly beneficial as reasoning complexity increases. Although \gls{auarc} decreases on HotpotQA despite improved \gls{auroc}, this likely reflects the substantially lower baseline accuracy of the benchmark, illustrating that discrimination quality and selective prediction utility capture different aspects of uncertainty estimation.

\begin{table}[tb]
\centering
\caption{\gls{ece} across benchmarks, per node and the \gls{bn} posterior (\textit{Overall}). Lower values indicate better calibration. Bold denotes the best ECE score in each row.}
\label{tab:ece}
\renewcommand{\arraystretch}{1.1}
\setlength{\tabcolsep}{5pt}
\begin{tabular}{llcccc}
\toprule
\textbf{Benchmark} & \textbf{Model}
    & \textit{Planner}
    & \textit{Evaluator}
    & \textit{Generator}
    & \textbf{BN (Ours)} \\
\midrule
\multirow{2}{*}{StrategyQA}
    & GPT-3.5-Turbo & 0.2186 & \textbf{0.2127} & 0.3703 & 0.4814 \\
    & GPT-4.1-Nano  & 0.1504 & \textbf{0.1205} & 0.4231 & 0.4942 \\
\midrule
\multirow{2}{*}{HotpotQA}
    & GPT-3.5-Turbo & 0.3410 & 0.4728 & 0.2502 & \textbf{0.1376} \\
    & GPT-4.1-Nano  & 0.3684 & 0.4199 & 0.2743 & \textbf{0.1120} \\
\bottomrule
\end{tabular}
\end{table}

The \gls{ece} results presented in Table \ref{tab:ece} indicate that calibration is strongly benchmark dependent. On StrategyQA, the Overall \gls{bn} substantially overestimates failure probability, suggesting that the deterministic OR gate propagates conservative uncertainty estimates even when the models answer correctly. In contrast, calibration improves considerably on HotpotQA, where the higher task difficulty aligns more closely with the \glspl{bn} uncertainty estimates. On HotpotQA, the Generator remains the strongest individual uncertainty signal, though its calibration is weaker than the Planner's and Evaluator's on StrategyQA. The calibration of the Evaluator node itself is inconsistent across benchmarks, reinforcing the Generator node uncertainty signal is the most reliable source at identifying failures, if not yet calibrated in absolute terms.

\begin{table}[tb]
\centering
\caption{Brier Score across benchmarks, per node and the \gls{bn} posterior (\textit{Overall}). Lower values indicate better probabilistic accuracy. The random baseline is 0.25; values below this indicate better-than-random calibration. Bold denotes the best Brier Score in each row.}
\label{tab:brier}
\renewcommand{\arraystretch}{1.1}
\setlength{\tabcolsep}{5pt}
\begin{tabular}{llccccc}
\toprule
\textbf{Benchmark} & \textbf{Model}
    & \textit{Planner}
    & \textit{Evaluator}
    & \textit{Generator}
    & \textbf{BN (Ours)} \\
\midrule
\multirow{2}{*}{StrategyQA}
    & GPT-3.5-Turbo & \textbf{0.2968} & 0.2989 & 0.3906 & 0.4831 \\
    & GPT-4.1-Nano & 0.2805 & \textbf{0.2635} & 0.4253 & 0.4816 \\
\midrule
\multirow{2}{*}{HotpotQA}
    & GPT-3.5-Turbo & 0.3197 & 0.4289 & 0.2781 & \textbf{0.2023} \\
    & GPT-4.1-Nano  & 0.3286 & 0.3745 & 0.2802 & \textbf{0.1788} \\
\bottomrule
\end{tabular}
\end{table}



The Brier Score results presented in Table \ref{tab:brier} reinforce the calibration trends observed in the \gls{ece} analysis. StrategyQA exhibits poor probabilistic calibration, reflecting a systematic mismatch between predicted and observed failure probabilities. In contrast, the Overall \gls{bn} achieves substantially improved Brier Scores on HotpotQA, demonstrating that its probability estimates become more accurate as task complexity increases. Together, the \gls{ece} and Brier Score results indicate that the proposed Bayesian uncertainty propagation framework produces better calibrated confidence estimates when the underlying uncertainty better reflects the true difficulty of the reasoning task.

\section{Evaluation}
\subsection{Does the Bayesian Network Add Value Over Individual Nodes?}

The evaluation demonstrates that the benefit of Bayesian uncertainty propagation depends on the reliability of the uncertainty signals being combined. On HotpotQA, the Overall \gls{bn} consistently outperforms the individual pipeline nodes, indicating that propagating uncertainty across multiple reasoning stages provides additional predictive value. In contrast, StrategyQA is dominated by the Generator uncertainty signal, suggesting that propagating weaker Planner and Evaluator signals can reduce overall performance and the use of P(True) self-evaluation techniques for the Generator node alone is sufficient for simpler queries to predict uncertainty. This behaviour highlights a limitation of the deterministic OR gate, which treats all uncertainty signals equally and cannot down-weight unreliable upstream nodes.

\subsection{Does Task Complexity Affect Uncertainty Propagation?}

The results indicate that task complexity has a substantial influence on the effectiveness of Bayesian uncertainty propagation. Compared with StrategyQA, HotpotQA exhibits stronger discrimination and calibration performance, suggesting that uncertainty estimates become more informative when reasoning requires multiple retrieval and reasoning stages. Although selective prediction performance decreases on HotpotQA, this is likely a consequence of the benchmark's substantially lower baseline accuracy rather than poorer uncertainty estimation. Overall, these findings suggest that uncertainty propagation provides the greatest benefit when uncertainty accumulates throughout multi-step reasoning.

\subsection{Calibration Versus Discrimination Trade-off}


The experiments demonstrate a clear trade-off between discrimination and calibration that depends on task difficulty. While the \gls{bn} provides stronger calibration on HotpotQA, StrategyQA reveals a tendency to overestimate failure probability, reflecting the conservative behaviour introduced by the deterministic OR gate. This conservatism is undesirable for well-calibrated probability estimation on simpler tasks but may be advantageous in safety-critical settings, where failing to identify an incorrect response is often more costly than issuing unnecessary warnings. Consequently, although the current framework requires improved calibration for general purpose applications, its risk-averse behaviour may be well suited to industrial decision support domains such as \gls{osw} maintenance scheduling.

\subsection{Limitations}
There are three principal limitations that constrain the generalisability of the current results. First is that the deterministic OR gate cannot adapt to benchmark-specific node reliability. As demonstrated in Table \ref{tab:auroc}, the Planner and Evaluator nodes show below-random \gls{auroc} scores on StrategyQA yet are weighted equally to the Generator in the OR gate, introducing noise into the Overall posterior and contributes to the below-UProp \gls{auroc} values on that benchmark (with GPT-4.1-Nano Overall \gls{auroc} as the exception). A noisy-OR parametrisation with learned inhibition parameters would address this issue by down-weighting nodes whose signals are empirically unreliable. This represents the most impactful architectural improvement available within the existing framework.

Secondly, the P(True) self-evaluation signal exhibits systematic conservatism, as the Generator assigns high uncertainty scores across a wide range of answers. This produces strong \gls{auroc} but poor Brier Score on StrategyQA. Using approaches such as post-hoc calibration with isotonic regression or Platt scaling fitted on a held-out validation set would allow the decoupling of the discriminatory quality from the calibration quality. This would be expected to drastically improve Brier Score on StrategyQA without degrading \gls{auroc} scores.

The third limitation is that the Wikipedia retrieval environment used in both benchmarks for this study differs from the knowledge bases that would be used in \gls{osw} maintenance scheduling applications. Uncertainty signals would behave differently in domain-specific settings and the current results should be interpreted as a proof-of-concept on publicly available multi-hop benchmarks rather than a direct validation of performance in the targeted industrial domain.

\section{Conclusion}

This paper presented a lightweight uncertainty-aware Agentic \gls{rag} framework in which per-node uncertainty signals derived from Wasserstein semantic divergence and P(True) self-evaluation are propagated through a \gls{bn} to estimate system-level confidence. The framework was evaluated on the StrategyQA and HotpotQA multi-hop reasoning benchmarks using GPT-3.5-Turbo and GPT-4.1-Nano across \gls{auroc}, \gls{auarc}, \gls{ece} and Brier Score. The proposed architecture provides a modular and computationally lightweight mechanism for monitoring uncertainty within multi-stage reasoning pipelines.

The experimental results demonstrate that Bayesian uncertainty propagation is most effective on tasks requiring genuine multi-hop reasoning, where uncertainty accumulates across multiple reasoning stages. While the Generator node's P(True) self-evaluation approach consistently provides the strongest individual uncertainty signal, combining uncertainty across pipeline stages improves overall performance on the more challenging benchmark. These findings support Bayesian uncertainty propagation as a promising approach for run-time uncertainty monitoring in Agentic \gls{rag} systems while also highlighting the importance of reliable upstream uncertainty estimates. 

The current framework remains limited by the deterministic OR gate, conservative P(True) calibration and evaluation on general purpose question-answering benchmarks rather than industrial data. Future work will investigate adaptive Bayesian structures, improved calibration techniques and validation using domain-specific retrieval corpora, including \gls{osw} maintenance decision support scenarios.

\section*{Data and Code Availability}

The datasets used in this article are publicly available and the code produced for this study, including the Agentic RAG pipeline, Bayesian Network inference scripts, and evaluation notebooks, is publicly available at:

\begin{center}
    \url{https://github.com/LouisDonaldson/BN-Uncertainty-Propagation.git}
\end{center}

\bibliography{refs}
\bibliographystyle{splncs04}
\end{document}